%% file: main.tex
\documentclass[journal]{IEEEtran}
\ifCLASSINFOpdf
\else
\fi

\newcommand{\ignore}[1]{}
\usepackage[boxed]{algorithm}
\usepackage{fancyhdr}
\usepackage[normalem]{ulem}
\usepackage[hyphens]{url}
\usepackage{microtype}
\usepackage{xcolor}
\usepackage{xspace}
\usepackage{multirow}
\usepackage{authblk}
\usepackage{array}

\newcommand{\eat}[1]{}

\newcommand{\tabincell}[2]{\begin{tabular}{@{}#1@{}}#2\end{tabular}}
\usepackage{array}
\makeatletter
\newcommand{\thickhline}{%
    \noalign {\ifnum 0=`}\fi \hrule height 1pt
    \futurelet \reserved@a \@xhline
}
\newcolumntype{"}{@{\hskip\tabcolsep\vrule width 1pt\hskip\tabcolsep}}

\usepackage{times}
\usepackage{graphicx}
\usepackage{latexsym}

\usepackage{amsmath}
\usepackage{amssymb}
\usepackage{booktabs}
\usepackage[colorlinks,
            linkcolor=black,
            anchorcolor=black,
            citecolor=black]{hyperref}

\begin{document}
%\title{Appearance Composing GAN: A General Method for Appearance-Controllable Human Video Motion Transfer}
\title{GAC-GAN: A General Method for Appearance-Controllable Human Video\\Motion Transfer}

\author{Dongxu~Wei,
        Xiaowei~Xu,
        Haibin~Shen,
        and~Kejie~Huang
        \thanks{D. Wei, H. Shen, and K. Huang are with the department of Information Science \& Electronic Engineering, Zhejiang University, Hangzhou 310027, China (Email: tracywei@zju.edu.cn; shen\_hb@zju.edu.cn; huangkejie@zju.edu.cn).}
        \thanks{X. Xiao is with Guangdong Provincial People’s Hospital, Guangdong Academy of Medical Sciences, Guangzhou, China, 510080. (Email: xiao.wei.xu@foxmail.com)}}% <-this % stops a space
\maketitle

\begin{abstract}
%Due to the rapid development of Generative Adversarial Networks (GANs), there has been significant progress in the field of human video motion transfer which has a wide range of applications in multimedia, computer vision and graphics.
Human video motion transfer has a wide range of applications in multimedia, computer vision and graphics. Recently, due to the rapid development of Generative Adversarial Networks (GANs), there has been significant progress in the field.
%However, almost all existing GAN-based works are prone to video synthesis where human motions are controllable while video appearances are not.
However, almost all existing GAN-based works are prone to address the mapping from human motions to video scenes, with scene appearances are encoded individually in the trained models. Therefore, each trained model can only generate videos with a specific scene appearance, new models are required to be trained to generate new appearances.
Besides, existing works lack the capability of appearance control. For example, users have to provide video records of wearing new clothes or performing in new backgrounds to enable clothes or background changing in their synthetic videos, which greatly limits the application flexibility.
%However, all the existing works are specific to the target video which is not general to other videos. Furthermore, they can only control human motions.
%Specifically, appearances of different video components (e.g. backgrounds and body part foregrounds) are bound together and not allowed to be altered alone.
%For example, the background and the foreground (e.g., head, upper body, lower body) cannot be changed.
%and can only occur at the same time. 
%Besides, such approach is video-specific and not generalized to other videos as it requires to train additional models to synthesize unseen video appearances.
%Obviously such approach is not flexible and efficient considering the large data in related applications. 
%approach is video-specific and can not generalized to other videos as it requires to train additional models to synthesize unseen video appearances.
In this paper, we propose GAC-GAN, a general method for appearance-controllable human video motion transfer.
To enable general-purpose appearance synthesis, we propose to include appearance information in the conditioning inputs. Thus, once trained, our model can generate new appearances by altering the input appearance information. To achieve appearance control, we first obtain the appearance-controllable conditioning inputs and then utilize a two-stage GAC-GAN to generate the corresponding appearance-controllable outputs, where we utilize an ACGAN loss and a shadow extraction module for output foreground and background appearance control respectively.
We further build a solo dance dataset containing a large number of dance videos for training and evaluation.
Experimental results show that, our proposed GAC-GAN can not only support appearance-controllable human video motion transfer but also achieve higher video quality than state-of-art methods.
%Example synthetic video results are available online: \url{https://youtu.be/8fhr5bcFM6Y}.
\end{abstract}


% \fi
\IEEEpeerreviewmaketitle
\input{introduction.tex}
\input{related_works.tex}
\input{method.tex}
\input{experiments.tex}
\input{conclusion.tex}
\bibliographystyle{IEEEtran}
\bibliography{ref}
\bibliographystyle{IEEEtran}
\input{ref.bbl}

\begin{IEEEbiography}[{\includegraphics[width=1in,height=1.25in,clip,keepaspectratio]{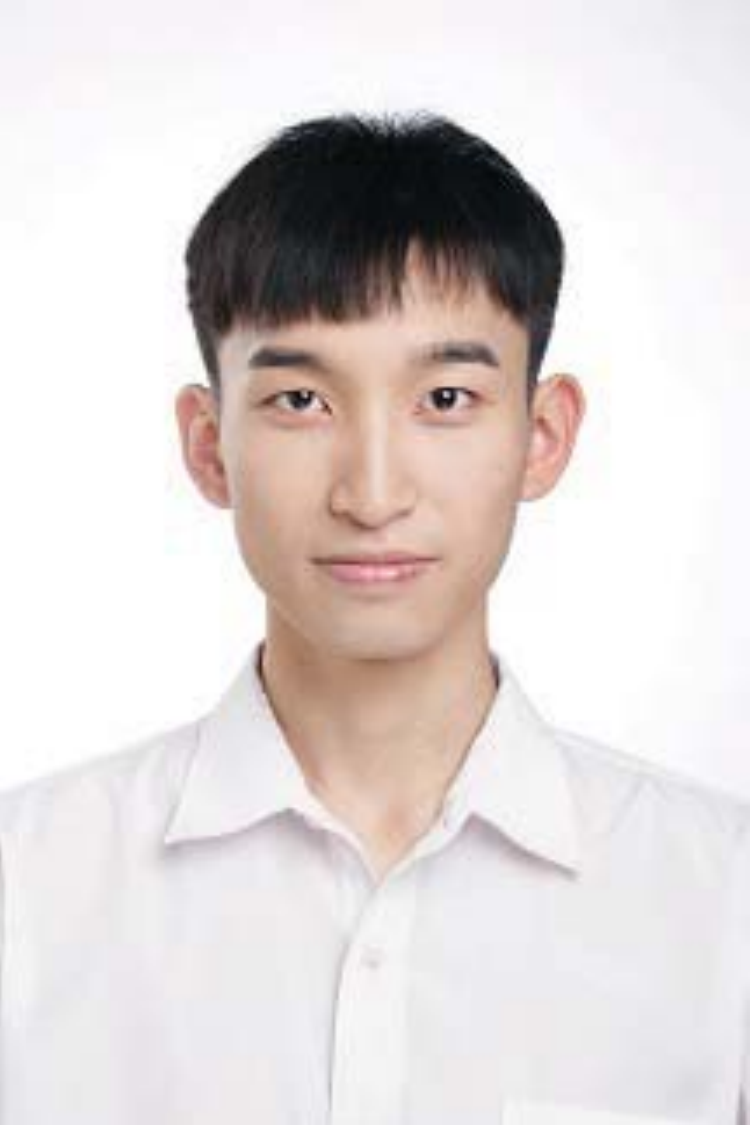}}]{Dongxu Wei (S'19)} received his B.S. degree in electronic science and technology from Harbin Institute of Technology. He is currently pursuing a Ph.D degree with the college of Information Science \& Electronic Engineering of Zhejiang University since 2017, advised by Prof. Haibin Shen. His research interests include image and video understanding, image and video synthesis, and computational photography.
\end{IEEEbiography}

\begin{IEEEbiography}[{\includegraphics[width=1in,height=1.25in,clip,keepaspectratio]{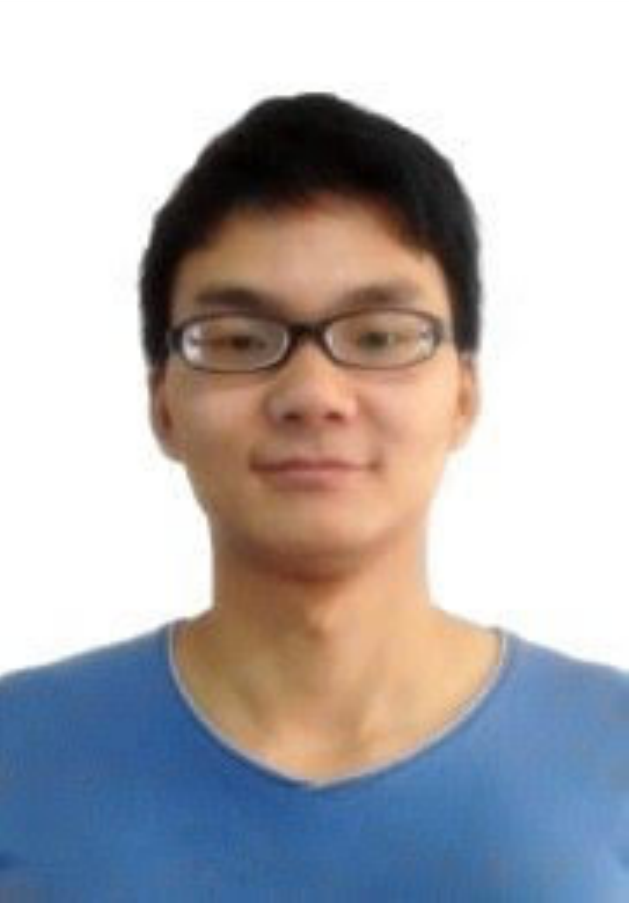}}]{Xiaowei Xu (S'14-M'17)}
received the B.S. and Ph.D. degrees in electronic science and technology from Huazhong University of Science and Technology, Wuhan, China, in 2011 and 2016 respectively.
He worked as a post-doc researcher at University of Notre Dame, IN, USA from 2016 to 2019.
He is now a AI researcher at Guangdong Provincial People's Hospital.
His research interests include deep learning, and medical image segmentation.
He was a recipient of DAC system design contest special service recognition reward in 2018
and outstanding contribution in reviewing, Integration, the VLSI journal in 2017.
He has served as TPC members in ICCD, ICCAD, ISVLSI and ISQED. He has published more than 50 papers in international peer-reviewed conference proceedings and journals.
\end{IEEEbiography}

\begin{IEEEbiography}[{\includegraphics[width=1in,height=1.25in,clip,keepaspectratio]{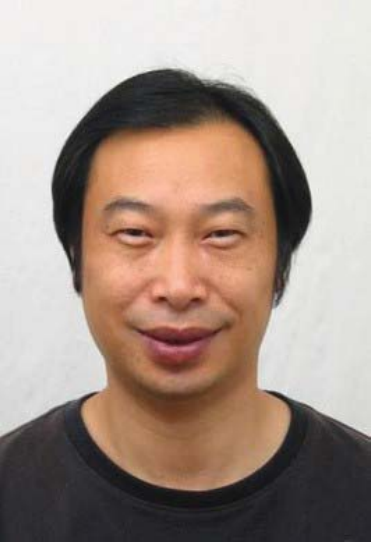}}]{Haibin Shen,} male, born in 1967, Ph.D., is a professor of Zhejiang University, the deputy director of HPEC (High Performance Embedded Computing) Key Lab of Ministry of Education and a member of the second level of 151 talents project of Zhejiang Province. His research interests are focused on learning algorithm, processor architecture and modeling. He has long-term working experience in system and IC both in university and industry. He has published over 30 SCI\&EI indexed papers in academic journals, and was granted more than 20 patents for inventions. He was also in charge of more than 10 national projects, including those from National Science and Technology Major Project and National High Technology Research and Development Project (863 Project), and participated in the projects of Chinese Nature Science Foundation and the formulation of national standards. His research achievement has been used by many authority organizations, and he was awarded the first prize of Electronic Information Science and Technology Award from Chinese Institute of Electronics.
\end{IEEEbiography}

\begin{IEEEbiography}[{\includegraphics[width=1in,height=1.25in,clip,keepaspectratio]{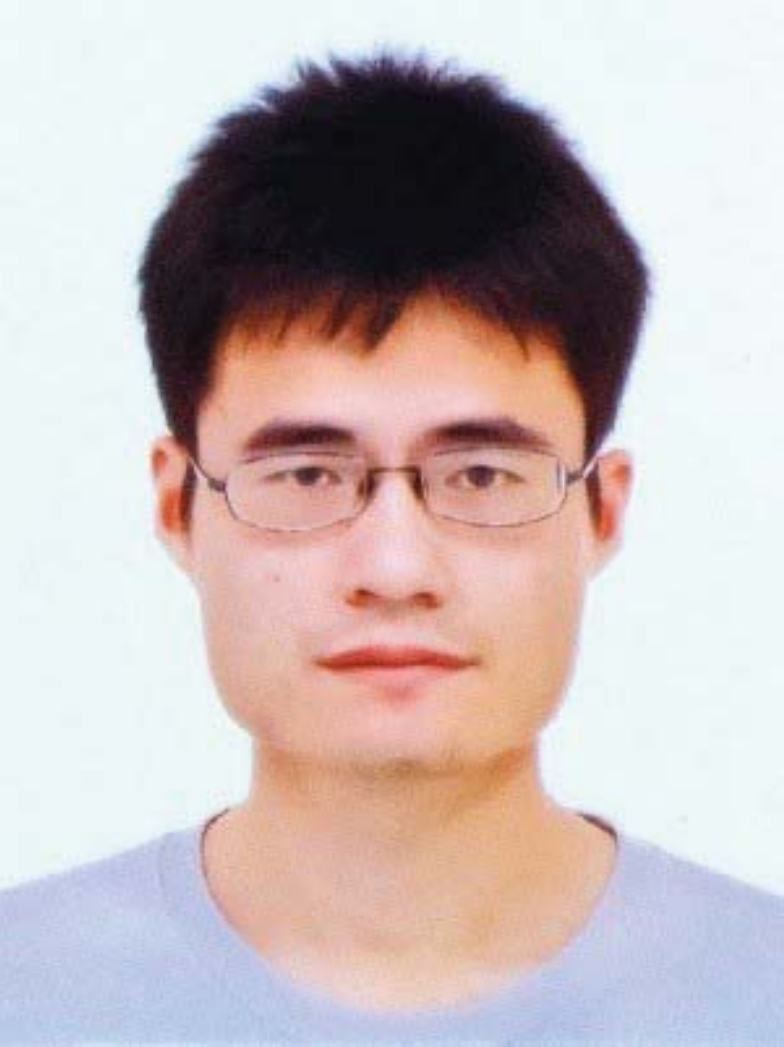}}]{Kejie Huang (M'13-SM'18)}
received his Ph.D degree from the Department of Electrical Engineering, National University of Singapore (NUS), Singapore, in 2014. He has been a principal investigator at College of Information Science \& Electronic Engineering, Zhejiang University (ZJU) since 2016. Prior to joining ZJU, he spent five years in the IC design industry including Samsung and Xilinx, two years in the Data Storage Institute, Agency for Science Technology and Research (A*STAR), and another three years in Singapore University of Technology and Design (SUTD), Singapore. He has authored or coauthored more than 30 scientific papers in international peer-reviewed journals and conference proceedings. He holds four granted international patents, and another eight pending ones. His research interests include low power circuits and systems design using emerging non-volatile memories, architecture and circuit optimization for reconfigurable computing systems and neuromorphic systems, machine learning, and deep learning chip design. He currently serves as the Associate Editor of the IEEE TRANSACTIONS ON CIRCUITS AND SYSTEMSPART II: EXPRESS BRIEFS.
\end{IEEEbiography}


%\end{IEEEbiography}


%\newpage



%\vfill


\end{document}

%% file: introduction.tex
\section{Introduction}
\IEEEPARstart{H}{uman} video motion transfer (HVMT) aims at synthesizing a video that the person in a target video imitates actions of the person in a source video, which is of great benefit to applications in scenarios such as games, movies and robotics. For example, the animation of virtual characters plays a key role in VR/AR games and movies. Based on HVMT techniques, we can animate the virtual game roles or movie actors freely to perform user-defined mimetic movements, thus rendering plausible visual results \cite{xu2018monoperfcap,shysheya2019textured}. Moreover, the animated visual data can be further utilized as simulated training data to train robotic agents that work for real-world situations, where real experiences may be hard to obtain \cite{dosovitskiy2017carla}.\par
With the recent emergence of Generative Adversarial Networks (GANs) \cite{goodfellow2014generative} and its variant conditional GANs (cGANs) \cite{mirza2014conditional}, there are many GAN-based works \cite{chan2019everybody,wang2018video,Liu:2019:NRR:3341165.3333002,zhou2019dance,aberman2019deep} that achieve great success in HVMT.
For ease of discussion, we decompose the video scene into scene appearance (background and human foreground) and human motion in the context of HVMT. 
Existing works have two limitations.
First, only the mapping from human motions to video scenes is addressed while scene appearances are encoded individually in the trained models. Therefore, once trained, each model is specific to the scene appearance of a target video and cannot generalize to other scene appearances. They have to train additional video-specific models with new target videos as the training data to generate new scene appearances.
%First, they are specific to the scene appearance of the target video (human and background appearances of the target video are encoded in the trained model), and thus cannot generalize to other scene appearances.
%In particular, they train their model with a specific target video and can only generate new videos with the same scene appearance (both human and background appearances) of the target video. They have to perform additional training with new target videos to generate new scene appearances.
Unfortunately, due to the large cost of manpower and computing resources produced by the data collection and the model training, such approach lacks efficiency for practical applications.
Second, existing methods can't control the scene appearance.
In particular, background and human foreground appearances are bound together and not allowed to be altered. Therefore, these methods can't synthesize videos with users wearing new clothes or performing in new backgrounds if users have never been in these clothes or backgrounds.
%In particular, background and human body part foregrounds (e.g., head, upper body and lower body) are bound together in the synthetic scene and not allowed to be altered alone. To synthesize videos with new clothes
%In fact, users expect to wear a variety of clothes without trying them on in backgrounds where they have not been [xxxx add reference here].
However, users expect to alter appearances in their synthetic videos without the efforts of real clothes and background changing.
%Thus, full control of motion, background, and human appearance including head, upper body and lower body can provide high flexibility in practical applications.\par
Thus, despite the human motion control, further appearance control is needed to provide high flexibility in practical applications.\par

In this work, we propose GAC-GAN: a general method for appearance-controllable human video motion transfer.
For general-purpose appearance synthesis, we propose to feed our model with appearance conditioning inputs in addition to motion conditioning inputs (body poses) used in other works, allowing the model to learn the mapping from human motions and scene appearances to video scenes.
%we propose to feed extracted human skeletons from videos to FAC-GAN, thus human appearances and background are not encoded in the model.
For appearance control, we propose to control output appearances through control of the conditioning inputs. Specifically, we propose a multi-source input selection strategy to first exert appearance control on the conditioning inputs during data preprocessing. Then a two-stage GAC-GAN framework which consists of a layout GAN and an appearance GAN is proposed to generate the corresponding appearance-controllable outputs from the conditioning inputs, where we further apply an elaborate ACGAN loss and a light-weight shadow extraction module to the appearance GAN to achieve control of the output human foreground and background respectively.
In our experiments, a large solo dance dataset including 148800 frames collected from 124 people is utilized for general-purpose training and evaluation. We first compare our approach against state-of-art video-specific \cite{wang2018video,chan2019everybody} and general-purpose \cite{balakrishnan2018synthesizing} methods through qualitative, quantitative and perceptual evaluations on the test set. The results show that, compared with other methods, our proposed approach can synthesize high-quality motion transfer videos that are perceptually more popular and quantitatively more similar to ground-truth real videos in a general way. Then we apply our method to ordinary and appearance-controllable HVMT tasks for further validation on simulated real-world situations where no ground-truth video is available. The results show that, in addition to the human motion control, our method can further control the appearances of the human foregrounds as well as the surrounding backgrounds flexibly. Moreover, to give a better insight into the proposed GAC-GAN framework, we conduct comprehensive ablation studies for our important components (i.e., multi-source input selection strategy, layout GAN, ACGAN loss and shadow extraction module).\par
To summarize, our main contributions are as follows:
\begin{itemize}
    \item We propose GAC-GAN: a general approach enabling appearance-controllable human video motion transfer.
    \item We achieve higher video quality than state-of-art methods by taking advantage of our novel component designs.
    \item We construct a large-scale solo dance dataset including a variety of solo dance videos for training and evaluation, which will be released publicly to facilitate future research.
\end{itemize}
The rest of the paper is structured as follows: Sec.\ref{related_work} discusses the related work. Sec.\ref{problem_form} introduces the problem formulation in our work. In Sec.\ref{method}, we describe the proposed GAC-GAN. In Sec.\ref{experiments}, we report and discuss our experimental results. Finally, Sec.\ref{conclusion} concludes the paper and discusses the future work.

%% file: related_works.tex
\section{Related Work}
\label{related_work}
\subsection{Classic motion transfer.}
Early works have attempted to reorder existing video frames \cite{efros2003recognizing,schodl2002controlled,schodl2000video} to obtain new videos consisting of frames with motions similar to the desired motions, where the results are not temporally coherent and can be easily distinguished from real videos. Later techniques try to animate coarse 3D character models \cite{hecker2008real,cheung2004markerless,lee1999hierarchical} to create rendered motion transfer videos, which results in coarse body silhouettes and unrealistic texture details. Recently, methods \cite{Pons2017ClothCap, Leroy2017Multi, 8100065} estimate detailed 3D characters with controllable body meshes to render plausible video results. However, most of these 3D rendering approaches require massive computation budgets dominated by the production-quality 3D reconstructions, which is inefficient for real-world applications.
\subsection{Image and video generation.}
Instead of relying on temporally incoherent video manipulations or computationally expensive 3D reconstructions, current motion transfer works depend more on image and video generation techniques. Traditional generation methods are prone to deal with syntheses of local textures based on simple hand-crafted features \cite{portilla2000parametric}. With the development of deep learning algorithms, variational autoencoder (VAE) \cite{kingma2013auto} and generative adversarial networks (GANs) \cite{goodfellow2014generative} become two mainstream methods due to their capabilities of synthesizing large-size images. Benefiting from the powerful two-player adversarial training, GAN-based generative models can synthesize images that are less blurry and more realistic than those generated by VAEs, which causes GANs are more exploited in image and video generation works. In the beginning, GAN-based image generation works \cite{10.5555/2969239.2969405, article} focus on designing GAN architectures to improve synthetic image resolutions. However, their image results are randomly generated from randomly sampled noises, which is out of user control. Since the emergence of conditional GANs (cGANs) \cite{mirza2014conditional}, works start to take class labels \cite{odena2017conditional, miyato2018cgans} or descriptive images \cite{isola2017image, zhu2017unpaired, liu2017unsupervised} as extra conditioning inputs to control the output image appearances, which belongs to the same method category as our proposed cGAN-based approach. Besides of image generation, there are also works \cite{vondrick2016generating, saito2017temporal, tulyakov2018mocogan, mathieu2016deep, liang2017dual, denton2017unsupervised} focus on synthesizing temporally coherent video sequences. For instance, unconditional video generation works \cite{vondrick2016generating, saito2017temporal, tulyakov2018mocogan} try to improve temporal consistency between adjacent synthetic frames based on GANs that consider not only visual quality but also temporal coherence. However, these approaches fail to generate high-quality or long-term video results, with scene appearances are randomly synthesized in an unconditional manner. Besides, video prediction techniques \cite{mathieu2016deep, liang2017dual, denton2017unsupervised} attempt to predict future video sequences based on the currently observed video sequences. Although the synthetic appearances are conditioned on the previous frames, future video motions are unconditionally generated, which is inappropriate for the motion controllable video synthesis that HVMT concerns.
%先介绍古老的image synthesis方法，非生成式的方法。然后介绍vae和gan两种生成式方法，表面vae会产生模糊，细节不如gan。然后说原始的gan只能生成随机的图像，则出现了cgan用来可控的图像生成。并且说到最开始利用cgan生成完整人体图像的方法，包括mocogan，TGAN，VGAN等，强调这些都是unconditional video synthesis或video prediction，人物面貌和动作是不可控的，与我们希望的每个关节动作都可以精确控制的人体图像生成不同。
\begin{figure*}
\centerline{\includegraphics[width=0.99\textwidth]{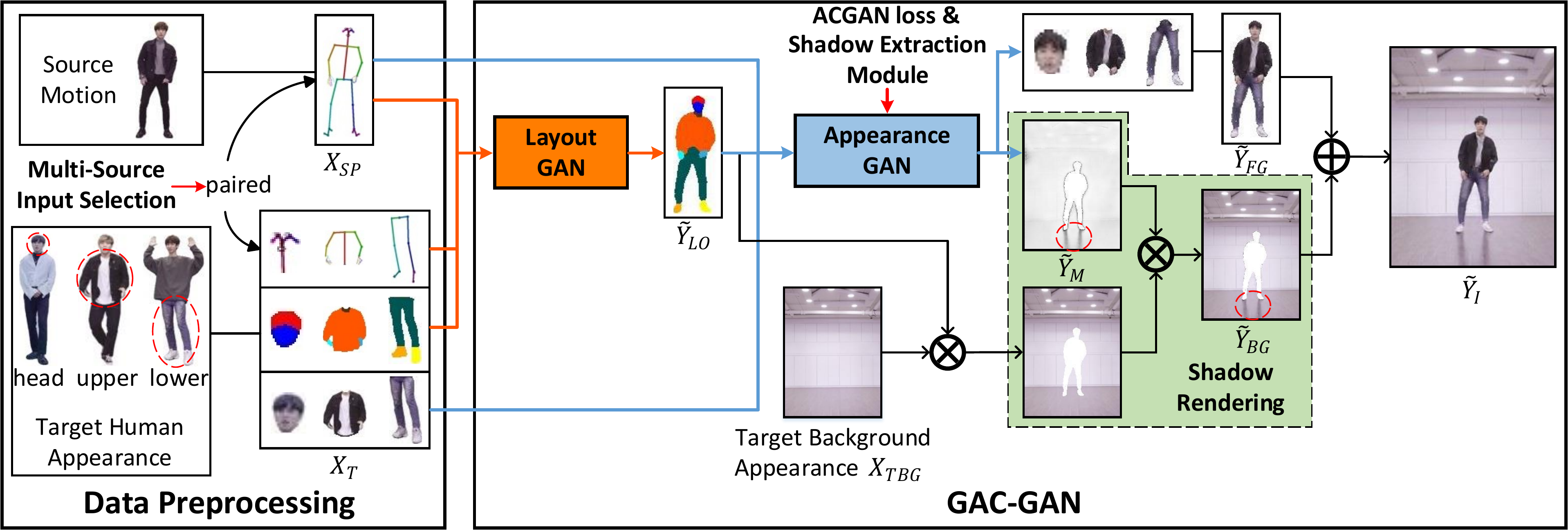}}
\caption{Overview of our method. In the data preprocessing, we obtain the paired $X_{SP}$ and $X_T$ from the source motion frame and the target appearance frames based on the multi-source input selection strategy. Then the processed inputs are fed into the GAC-GAN which consists of a layout GAN and an appearance GAN to sequentially generate the layout $\tilde{Y}_{LO}$ and the scene appearance $\tilde{Y}_{I}$ (composed of the synthetic foreground $\tilde{Y}_{FG}$ and the rendered background $\tilde{Y}_{BG}$), where we further apply an ACGAN loss and a shadow extraction module to the appearance GAN to control foreground and background appearances respectively. In the figure, the orange and the blue arrows represent data flows of the layout GAN and the appearance GAN respectively, $\otimes$ and $\oplus$ represent pixel-wise multiplication and addition operations respectively. In the data preprocessing module, the red circles specify the desired body part appearances. In the GAC-GAN module, the red circles point out how the synthetic shadow map $\tilde{Y}_{M}$ modulates brightness for the input background image $X_{TBG}$, which enables shadow rendering.} \label{fig1}
\end{figure*}
\subsection{GAN-based motion transfer.}
Due to the great success of the GAN-based image and video generation approaches mentioned above, many works are developed for motion transfer based on them.
\subsubsection{Image-based human pose transfer}
In the recent years, there have been significant efforts which we refer to as image-based methods \cite{balakrishnan2018synthesizing,neverova2018dense,zanfir2018human,siarohin2018deformable,dong2018soft,liang2019pcgan,wang2019example} aiming at synthesizing new pose images given the human appearance of a single input image. The purpose of these image-based works is to impose the input human appearance onto new poses in an image-to-image translation manner \cite{isola2017image}, which is very similar to the human video motion transfer that we focus on. \cite{balakrishnan2018synthesizing,neverova2018dense,zanfir2018human} utilize spatial transformations or surface deformations to transform the input appearance texture into new pose layouts, where the transformed results are rough and refined in detail to generate output images. Similarly, \cite{siarohin2018deformable,dong2018soft,liang2019pcgan} apply such transformations or deformations to appearance features instead of textures, where the transformed features are then decoded to generate new pose images. Furthermore, \cite{wang2019example} propose a style discriminator to force the generator to preserve the input appearance style, which gives a new sight from the aspect of discriminator design. Although these image-based methods can achieve general-purpose appearance synthesis, all of them are designed for still image generation without consideration of temporal coherence, which causes they are not qualified for video synthesis that we concern. Besides, these methods try to generate unseen body views from a single input image, which greatly restricts their performance due to the lack of appearance information, especially when the desired output pose greatly differs from the input pose.
\subsubsection{Video-based human motion transfer}
As the video counterpart of the above mentioned image-based pose transfer, video-based motion transfer considers video generation with access to more appearance information contained in a whole video, leading to a higher level of temporal coherence and visual quality.
In \cite{wang2018video}, the authors propose to generate optical flows to warp previously generated frames into temporally consistent new frames. Besides, \cite{chan2019everybody} use a temporal smoothing loss to enforce temporal consistency between adjacent frames. Note that video quality depends not only on temporal coherence but also on appearance details.
Thus recent works come up with feeding rendered images of 3D models \cite{Liu:2019:NRR:3341165.3333002} or transformed images of body parts \cite{zhou2019dance} into their models as input conditions to obtain realistic appearances. Moreover, \cite{aberman2019deep} split their network into two training branches with respect to appearance generation and temporal coherence improvement to account for both sides.
Although these works can generate videos with higher quality than image-based methods, an obvious limitation is that they have to train additional models to generate unseen scene appearances, keeping them from general-purpose appearance synthesis required in real-world applications. Besides, none of them can realize controllable appearance synthesis to satisfy user demands for clothes and background changing. Although \cite{zhou2019dance} can support background replacement with user-defined images, they don't allow users to try on different clothes in the synthetic videos.
%In our work, a new video-based approach is proposed, allowing for appearance-controllable video synthesis as well as straightforward generalization to different human subjects.

%% file: method.tex
\begin{figure*}
\centerline{\includegraphics[width=0.99\textwidth]{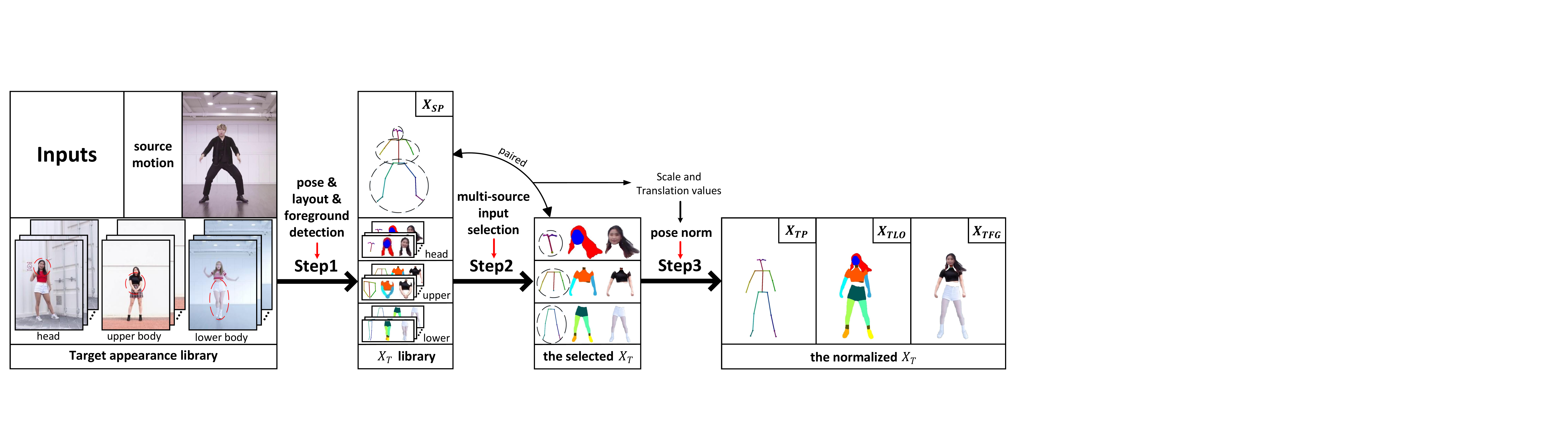}}
\caption{Illustration of our data preprocessing. In step 1, we detect poses, layouts and foregrounds to obtain the motion condition $X_{SP}$ and the appearance condition ($X_T$) library, where each red circle specifies a target body part. In step 2, each body part of the source pose $X_{SP}$ is paired with a target body part (pose $X_{TP}$, layout $X_{TLO}$, and foreground $X_{TFG}$) in the $X_T$ library according to body part pose similarity. Then in step 3, we use the computed scale and translation values between body parts of $X_{SP}$ and $X_{T}$ to transform the body parts of $X_T$ into the same sizes and positions as those of $X_{SP}$.}
\label{fig2}
\end{figure*}
\section{Problem Formulation}
\label{problem_form}
Before describing our method, we first define the problem to solve: given conditioning input of a source motion video and multiple target appearance videos, we aim at synthesizing a new video with human motion of the source video and combined scene appearance of the target videos. Specifically, the conditioning input is divided into motion conditioning input (source motion) and appearance conditioning input (target appearance), where the target appearance is further divided into human and background appearances. Source motion input is described by the estimated body poses of the source video frames. To enable human appearance control, target human appearance input is decomposed into three user-defined body parts (e.g., head, upper body and lower body) with respect to appearances of face, upper garment and lower garment, each of which is described by the estimated body part poses, layouts and foregrounds of its own target video frames. To enable background appearance control, target background appearance input is described by a user-defined background image. Conditioned on the source motion and the target appearance inputs, we generate the corresponding outputs including body layouts, body part foregrounds and shadow maps, where the body layouts are generated by the layout GAN while the others are generated by the appearance GAN. Then we use the synthetic shadow maps to render shadows on the input background image. Finally, we obtain the synthetic full scenes by composing the synthetic body part foregrounds and the rendered backgrounds together. For the above mentioned inputs and outputs, we give their variable definitions used in this paper as follows:
\begin{enumerate}
\item{\textbf{Inputs}}
\begin{itemize}
    \item {\textbf{source motion}: source pose $X_{SP}$}
    \item {\textbf{target human appearance} ($X_T$):\\
    target poses: $X_{TP,H}$, $X_{TP,U}$, $X_{TP,L}$\\
    target layouts: $X_{TLO,H}$, $X_{TLO,U}$, $X_{TLO,L}$\\
    target foregrounds: $X_{TFG,H}$, $X_{TFG,U}$, $X_{TFG,L}$}
    \item {\textbf{target background appearance}: $X_{TBG}$}
\end{itemize}
\item{\textbf{Outputs}}
\begin{itemize}
    \item {\textbf{layout GAN}: body layout $\tilde{Y}_{LO}$}
    \item {\textbf{appearance GAN}:\\
    body part foregrounds: $\tilde{Y}_{FG,H}$, $\tilde{Y}_{FG,U}$, $\tilde{Y}_{FG,L}$\\
    shadow map: $\tilde{Y}_{M}$\\
    background: $\tilde{Y}_{BG}$\\
    full scene: $\tilde{Y}_{I}$}
\end{itemize}
\end{enumerate}where $X$ and $\tilde{Y}$ mean input and output, $S$ and $T$ represent input source and target videos, $P$, $LO$, $FG$, $BG$, $M$, $I$ represent pose, layout, foreground, background, shadow map and scene image, $H$, $U$, $L$ refer to head, upper body and lower body.
\section{Method}
\label{method}
In this section, we first give the overview of our proposed method, which is followed by two subsections with respect to our data preprocessing and GAC-GAN framework.
\subsection{Overview}
The overview of our method is depicted in Figure \ref{fig1}.\par
First, we apply \emph{\textbf{data preprocessing}} to the input videos to obtain our conditioning inputs, where we pair each motion conditioning input (source pose $X_{SP}$) with an optimal appearance conditioning input $X_T$ (target pose $X_{TP}$, layout $X_{TLO}$ and foreground $X_{TFG}$ of head, upper body and lower body) based on a \textbf{multi-source input selection strategy}. Specifically, each body part of $X_T$ is obtained from its own target human appearance source, which can be altered based on user preferences to enable input appearance control.\par
%Specifically, the proposed strategy selects appearance conditions for different body parts (e.g., head, upper and lower body) separately, which enables full appearance control on inputs by selecting appearance conditions of different parts from different target video sources (multi-source). Since each body part appearance condition is selected based on its pose similarity with the corresponding body part of the desired source pose, we can ensure that the obtained appearance inputs contain the maximum appearance information needed for appearance synthesis. Moreover, given multi-source appearance inputs are not available for real videos in our training data, selecting different body part appearance conditions from different video frames can eliminate the difference between multi-source inputs used in testing and single-source inputs used in training.\par
Next, we feed the motion ($X_{SP})$ and the appearance ($X_T$) conditioning inputs into our two-stage \emph{\textbf{GAC-GAN}} which consists of a layout GAN and an appearance GAN, responsible for controllable layout synthesis and appearance synthesis respectively. Because generating appearances directly from body pose points can be extremely hard, we can ease training by dividing our model into these two stages, where the synthetic layout can be regarded as the intermediate representation of the final appearance result. Specifically, in the first stage, the \textbf{layout GAN} is designed to synthesize the foreground layout $\tilde{Y}_{LO}$ whose body pose and body part distribution are consistent with the motion condition ($X_{SP})$ and the multi-source appearance condition ($X_{TLO}$) respectively.
In the second stage, the \textbf{appearance GAN} takes the synthetic layout $\tilde{Y}_{LO}$ as additional motion conditioning input to generate the desired scene appearance $\tilde{Y}_{I}$, which is composed of a synthetic foreground $\tilde{Y}_{FG}$ and a rendered background $\tilde{Y}_{BG}$. As for the foreground, we train the appearance GAN with an \textbf{ACGAN loss} to ensure the appearance consistency between the synthetic foreground and the input appearance condition, which therefore enables foreground appearance control in consistency with the input appearance control. As for the background, we implant a light-weight \textbf{shadow extraction module} into the appearance GAN to generate a shadow map $\tilde{Y}_{M}$ that modulates background brightness and renders appearance-irrelevant shadows on ${X}_{TBG}$, which therefore enables background appearance control by directly replacing background with arbitrary user-defined images.
\subsection{Data Preprocessing}
The main purpose of data preprocessing is to obtain our motion and appearance conditioning inputs. For each frame synthesis, the motion condition is extracted from a source motion frame while the appearance condition is extracted from a target appearance library which contains three kinds of target appearance video frames with respect to head, upper body and lower body. Since video sources of the three body parts are alterable based on user preferences, the multi-source input appearance condition is fully appearance-controllable. With body motion is specified by the motion condition, the data preprocessing aims at obtaining the paired appearance condition which contains the maximum appearance information needed for appearance synthesis. Specifically, the data preprocessing consists of the three steps depicted in Figure \ref{fig2}, where the multi-source input selection strategy utilized in step 2 is the key to ensure the obtained appearance condition is optimal for the motion condition. It's noted that there's no restriction on the frame number for the target appearance library, we can obtain the optimal appearance condition no matter how many frames are provided.
%The main purpose of data preprocessing is to obtain the optimal appearance conditioning input ($X_T$) with respect to the desired source motion ($X_{SP}$), where each body part has its own appearance inputs selected from the corresponding target videos to enable separate control on inputs of different body parts. The preprocessing consists of the following three steps, aiming at obtaining the maximum available input appearance information for appearance synthesis of the desired body motions ($X_{SP}$). It's noted that there's no restriction on the input frame number, we can obtain optimal appearance inputs no matter how many frames are provided.
\subsubsection{Step 1: Detecting Poses, Layouts and Foregrounds}
\begin{figure}
\centerline{\includegraphics[width=0.46\textwidth]{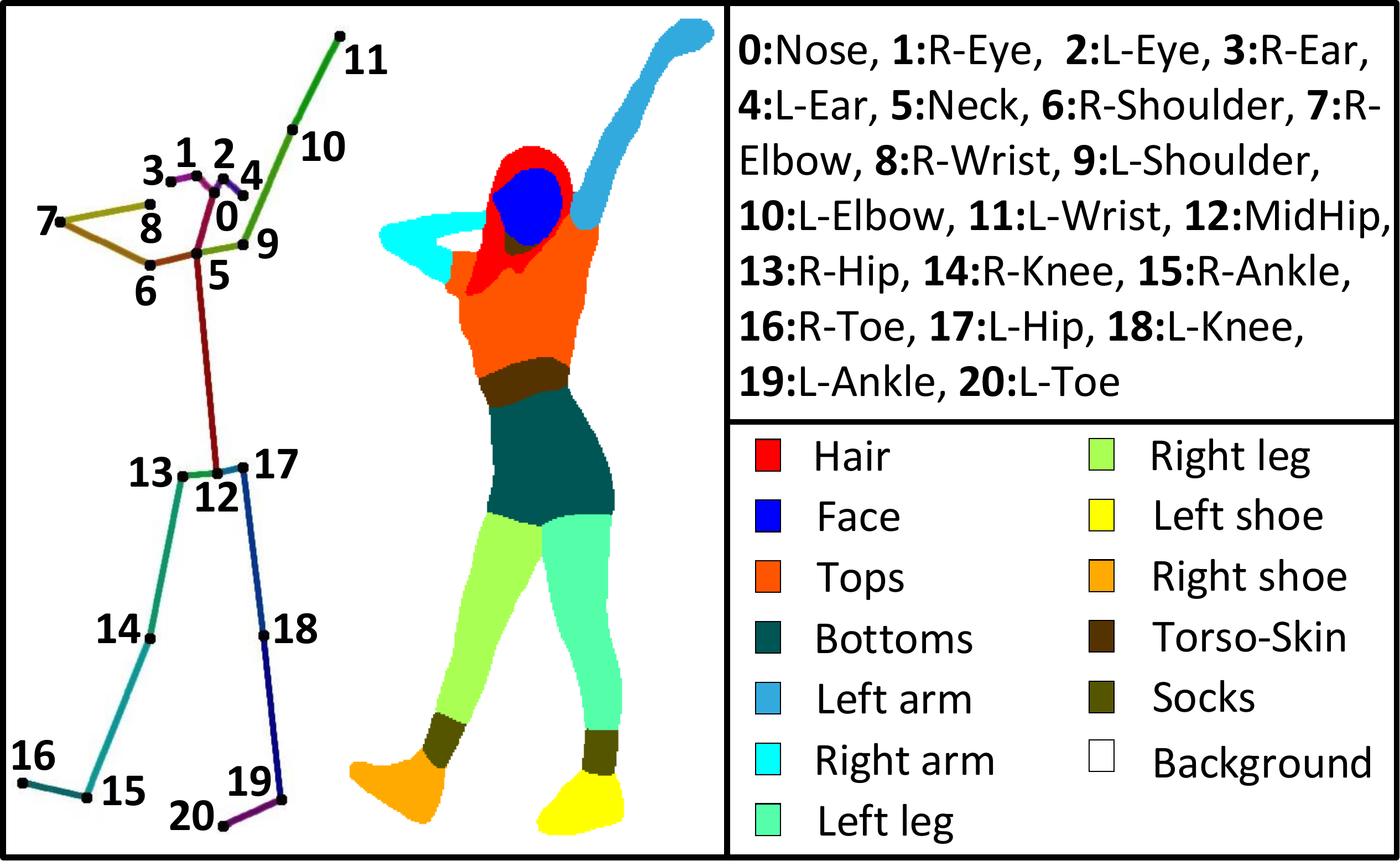}}
\caption{Illustration of pose and layout detection results. Pose points and semantic labels are distinguished by numbers and colors respectively. "R-" means "right" and "L-" means "left".}
\label{fig3}
\end{figure}

We utilize \cite{cao2017realtime} and \cite{gong2018instance} to detect body poses and semantic layouts respectively, where the pose point locations and the layout classes are described in Figure \ref{fig3}. Then we can decompose the full body layouts into body part layouts for the three body part regions. Specifically, head region is a combination of hair and face; upper body region is a combination of tops, torso skin, left arm and right arm; lower body region is a combination of bottoms, left leg, right leg, left shoe, right shoe and socks. Thereafter we can extract foregrounds for each body part by multiplying the full images with the corresponding body part masks derived from the body part layouts. Based on the detections described above, we can obtain the input motion condition from the source motion frame and the appearance condition library from the target appearance library. In particular, the motion condition is the detected source body pose ($X_{SP}$) of the input source motion frame. The appearance condition library ($X_T$ library) consists of three body part appearance condition libraries, each of which contains target body part poses ($X_{TP}$), layouts ($X_{TLO}$) and foregrounds ($X_{TFG}$) obtained from the corresponding video of the target appearance library.
\subsubsection{Step 2: Multi-Source Input Selection}
Since human appearances vary significantly with body poses, we propose a multi-source input selection strategy to select the optimal $X_T$ based on the body part pose similarity. For each body part of the motion condition ($X_{SP}$), we select the paired body part appearance condition from the corresponding body part appearance condition library, where the pose of the selected body part appearance condition is the most similar to the body part pose of the motion condition within the body part appearance condition library. Thus we obtain the selected $X_T$ which consists of three body part appearance conditions. Each body part appearance condition is composed of a body part pose, layout and foreground, containing the maximum appearance information needed for body part appearance synthesis.
Specifically, pose similarity of each body part is denoted as the average cosine similarity between the corresponding source and target body part pose vectors:
\begin{small}
\begin{equation}
\begin{aligned}
Sim=\frac{1}{N}\sum_{i=1}^{N}\frac{\overrightarrow{V_S^i}\cdot\overrightarrow{V_T^i}}{\vert{\overrightarrow{V_S^i}}\vert\,\vert{\overrightarrow{V_T^i}}\vert}
\label{con:E1}
\end{aligned}
\end{equation}
\end{small}where $Sim$ is the body part pose similarity, $\overrightarrow{V_S^i}$ and $\overrightarrow{V_T^i}$ represent the i-th body part pose vectors of the source pose $X_{SP}$ and the target pose $X_{TP}$ respectively, $|\overrightarrow{V_S^i}|$ and $|\overrightarrow{V_T^i}|$ represent vector lengths of $\overrightarrow{V_S^i}$ and $\overrightarrow{V_T^i}$ respectively. $N$ is the number of body part pose vectors, which equals to 5, 7, 8 for pose vectors of head, upper body, lower body.
In particular, head pose vectors are $\overrightarrow{{P_0}{\,}{P_1}}$, $\overrightarrow{{P_0}{\,}{P_2}}$, $\overrightarrow{{P_1}{\,}{P_3}}$, $\overrightarrow{{P_2}{\,}{P_4}}$, $\overrightarrow{{P_0}{\,}{P_5}}$. Upper body pose vectors are $\overrightarrow{{P_5}{\,}{P_6}}$, $\overrightarrow{{P_6}{\,}{P_7}}$, $\overrightarrow{{P_7}{\,}{P_8}}$, $\overrightarrow{{P_5}{\,}{P_9}}$, $\overrightarrow{{P_9}{\,}{P_{10}}}$, $\overrightarrow{{P_{10}}{\,}{P_{11}}}$, $\overrightarrow{{P_5}{\,}{P_{12}}}$. Lower body pose vectors are $\overrightarrow{{P_{12}}{\,}{P_{13}}}$, $\overrightarrow{{P_{13}}{\,}{P_{14}}}$, $\overrightarrow{{P_{14}}{\,}{P_{15}}}$, $\overrightarrow{{P_{15}}{\,}{P_{16}}}$, $\overrightarrow{{P_{12}}{\,}{P_{17}}}$, $\overrightarrow{{P_{17}}{\,}{P_{18}}}$, $\overrightarrow{{P_{18}}{\,}{P_{19}}}$, $\overrightarrow{{P_{19}}{\,}{P_{20}}}$.
In the above description, $P_0{\sim}P_{20}$ represent pose points marked as numbers as shown in Figure \ref{fig3}.
\subsubsection{Step 3: Pose Normalization}
Although body parts of the selected $X_T$ have the most similar poses with those of $X_{SP}$, sizes and positions of different parts are not compatible with each other and therefore needed to be normalized to form a whole body spatially consistent with $X_{SP}$. In practice, we apply a pose normalization to transform each body part of $X_T$ into the same size and position as the corresponding part of $X_{SP}$, where the scale values and the translation distances of different parts are computed separately by analyzing the differences between body parts of $X_{SP}$ and $X_{TP}$ in vector lengths and point locations:
\begin{small}
\begin{equation}
\begin{aligned}
Scale&=\frac{\sum_{i=1}^{N_v}\vert\overrightarrow{V_S^i}\vert}{\sum_{i=1}^{N_v}\vert\overrightarrow{V_T^i}\vert}\\
Translation&=\frac{1}{N_p}\sum_{j=1}^{N_p}(P_S^j-P_T^j)
\label{con:E2}
\end{aligned}
\end{equation}
\end{small}where $N_v$ is the number of body part pose vectors, $N_p$ is the number of body part pose points, $P_{S}$ and $P_{T}$ represent source and target body part pose points respectively. For head, $N_p=6$, pose points are $P_0{\sim}P_5$. For upper body, $N_p=8$, pose points are $P_5{\sim}P_{12}$. For lower body, $N_p=9$, pose points are $P_{12}{\sim}P_{20}$.
\par
Thus we obtain the transformed body part pose points, layouts and foregrounds. Then the pose points of different parts are connected to compose a new target pose $X_{TP}$ while the body part layouts are processed into a one-hot tensor $X_{TLO}$ with each channel represents a body part as shown in Figure \ref{fig3}. Similarly, the body part foregrounds are also processed into a tensor $X_{TFG}$ which consists of body part channels consistent with $X_{TLO}$. By separating different body parts by different channels, we can eliminate the loss of appearance information caused by the overlap between body parts that come from different video frames. Moreover, since the obtained body parts are inherently misaligned, we can eliminate the difference between single-source and multi-source appearance inputs, which benefits our training because only single-source inputs are available during training due to the lack of ground truths for multi-source appearance outputs.
\begin{figure*}
\centerline{\includegraphics[width=0.9\textwidth]{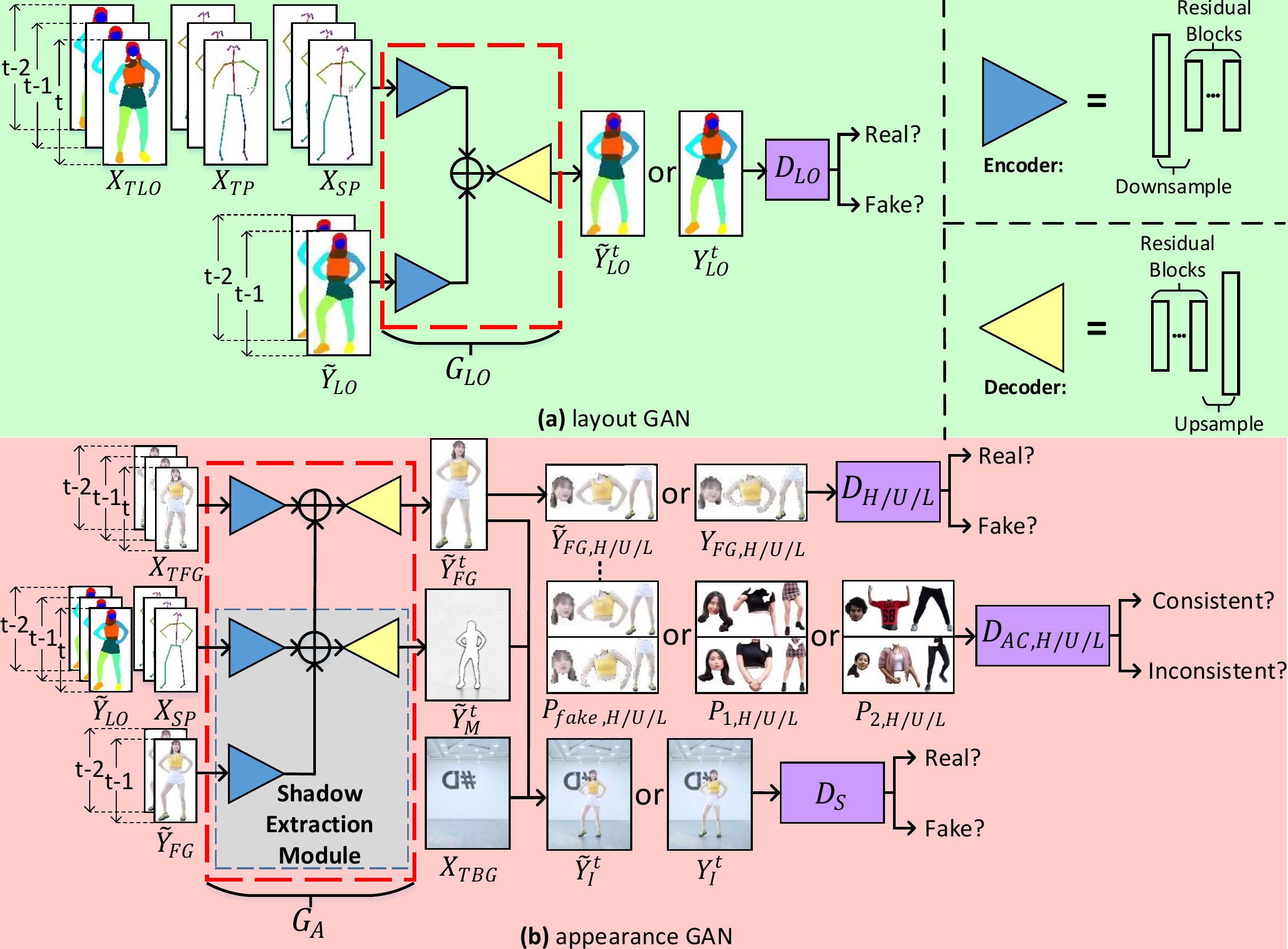}}
\caption{Illustration of the GAC-GAN. (a) and (b) depict frameworks of the layout GAN and the appearance GAN respectively, where encoder and decoder architectures are also drawn above. In (b), foregrounds and discriminators of the three body parts are drawn in the same blocks annotated by ${H/U/L}$ for simplicity, which are separated in practice.} \label{fig4}
\end{figure*}
\subsection{GAC-GAN}
Given appearance is fully controllable in the conditioning input, the GAC-GAN is designed to generate the corresponding fully controllable appearance output.
As shown in Figure \ref{fig4}, our GAC-GAN has two stages: a layout GAN and an appearance GAN, described in detail in the following subsections.
It's noted that, because videos are generated frame by frame, we present the generation of the frame at time $t$ as an example in the following discussions for convenience.
\subsubsection{Layout GAN}
The layout GAN aims at synthesizing the desired multi-source body layout with body part distributions consistent with the multi-source appearance condition. By taking the synthetic layout as additional motion condition, we can describe the human motion at a more accurate pixel level compared to other works \cite{chan2019everybody,wang2018video,Liu:2019:NRR:3341165.3333002,zhou2019dance,aberman2019deep} that use sparse body pose points as motion conditions.\par
\emph{Network Architectures:}
Our layout GAN is made of a layout generator $G_{LO}$ and a layout discriminator $D_{LO}$ as shown in Figure \ref{fig4}(a). Specifically, the generator $G_{LO}$ consists of two encoders and one decoder. The first encoder learns to encode features for the concatenation of three consecutive source poses, target poses and target layouts: $X_{LO}\vert^t_{t-2}=[X_{SP}\vert^t_{t-2},X_{TP}\vert^t_{t-2},X_{TLO}\vert^t_{t-2}]$. The second encoder learns to encode features for the concatenation of two previously generated layouts: $\tilde{Y}_{LO}\vert^{t-1}_{t-2}$. Then the two kinds of features are summed and fed into the decoder to generate the desired layout $\tilde{Y}_{LO}^t$. Here we include features of the concatenated consecutive frames to improve temporal consistency. Besides, the discriminator $D_{LO}$ is designed to be multi-scale \cite{isola2017image} to determine whether the generated layout is real or fake.\par
\emph{Objective Function:}
To train the layout GAN, we design the objective like this:
\begin{small}
\begin{equation}
\begin{aligned}
L_{LO}=L_{GAN}^{LO}+\lambda_{SS}L_{SS}^{LO}+\lambda_{T}L_{T}^{LO}+\lambda_{FM}L_{FM}^{LO}
\label{con:E3}
\end{aligned}
\end{equation}
\end{small}$L_{GAN}^{LO}$ is the adversarial loss of the layout GAN, which is given by:
\begin{small}
\begin{equation}
\begin{aligned}
L_{GAN}^{LO}=&E[log D_{LO}(Y_{LO},X_{LO})\\&+log[1-D_{LO}(\tilde{Y}_{LO},X_{LO})]]
\label{con:E4}
\end{aligned}
\end{equation}
\end{small}where $Y_{LO}$ is the real layout map with respect to $\tilde{Y}_{LO}$.\\
$L_{SS}^{LO}$ is the structural sensitive loss adapted from \cite{liang2018look} and weighted by $\lambda_{SS}$, which is used to minimize the difference between $Y_{LO}$ and $\tilde{Y}_{LO}$ at both the pixel level and the structure level. It can be derived like this:
\begin{small}
	\begin{equation}
	\begin{aligned}
	&L_{SS}^{LO}=L_{joint} \cdot L_{pixel},\\
	&L_{joint}=\frac{1}{2n}\sum_{i=1}^{n}{\Vert C_{i,real}-C_{i,fake}\Vert}_2^2
	\label{con:E5}
	\end{aligned}
	\end{equation}
\end{small}where the pixel-wise softmax loss $L_{pixel}$ is weighted by the joint structure loss $L_{joint}$, which is an L2 loss used to measure the structural difference between the real and the generated layout maps. $C_{i,real}$ and $C_{i,fake}$ represent center points of the real and the generated layout maps, respectively, which are computed by averaging coordinate values of the i-th layout regions for the two layout maps. Specifically, when $i$ ranges from $1$ to $n$ ($n=9$), the i-th region represents: head, tops, bottoms, left arm, right arm, left leg, right leg, left shoe and right shoe. As shown in Figure \ref{fig3}, all the regions have their class labels except for the head, which is a merged region of face and hair.\\
$L_{T}^{LO}$ weighted by $\lambda_{T}$ is the temporal loss used to minimize temporal difference between the real and the generated layout sequences, which can be derived like this:
\begin{small}
	\begin{equation}
	\begin{aligned}
	L_{T}^{LO}=E[log D_{LO}^{T}({S_{LO}})+log[1-D_{LO}^{T}(\tilde{S}_{LO})]]
	\label{con:E6}
	\end{aligned}
	\end{equation}
\end{small}where $D_{LO}^{T}$ is the temporal discriminator of the layout GAN, trained to determine whether a layout sequence is real or fake. $S_{LO}$ and $\tilde{S}_{LO}$ are the real and the generated layout sequences, which are obtained by concatenating three consecutive ${Y}_{LO}$s and $\tilde{Y}_{LO}$s sampled by the sampling operator presented in vid2vid \cite{wang2018video}.\\
$L_{FM}^{LO}$ is the discriminator feature matching loss presented in pix2pixHD \cite{wang2018high} and weighted by $\lambda_{FM}$, which is used to improve synthesis quality.
\subsubsection{Appearance GAN}
Provided with the additional synthetic motion condition that specifies the desired body layout, the appearance GAN aims at synthesizing the desired foreground and background appearances, which are added together to compose the full scene appearance.\par
As for the foreground, since the appearance is already controllable in the input appearance condition, we can synthesize the corresponding controllable foreground appearance by ensuring the appearance consistency between the synthetic and the input appearances. Therefore, we propose an \textbf{ACGAN loss} to supervise not only visual quality but also appearance consistency during training. Besides, since ground truths for multi-source appearance outputs don't exist, we utilize three part-specific ACGAN losses with respect to head, upper body and lower body to supervise different body parts separately rather than supervise them as a whole, which helps to alleviate inner relevance between body parts that come from the same videos in our training data.\par
As for the background, we implant a light-weight \textbf{shadow extraction module} into the appearance GAN to generate the shadow map that modulates background brightness and renders background shadow rather than directly generate the background appearance from scratch \cite{chan2019everybody,wang2018video,Liu:2019:NRR:3341165.3333002,zhou2019dance,aberman2019deep}. The reasons are manifold: 1) Since video backgrounds are fixed and can be regarded as still images, patterns of the backgrounds are much fewer than those of the foregrounds in the training data. A deep learning model could easily get overfitted if trained with a few kinds of background appearances for a large number of training steps. 2) Besides, an overfitted model may tend to remember the relevance between co-occurred foreground and background appearances, which may cause failures when synthesizing new human foregrounds. 3) Compared to generating the fixed background appearance which can be easily described by a still image, generation of the appearance-irrelevant background shadow is more worth studying, which enables background appearance control by adding shadows to alterable user-defined background images.\par
Then we describe the architectures in detail to explain the above mentioned functionalities. As shown in Figure \ref{fig4}(b), the appearance GAN is made of an appearance generator $G_{A}$, a scene discriminator $D_{S}$, three standard body part discriminators $D_{H}$, $D_{U}$, $D_{L}$ and three appearance-consistency body part discriminators $D_{AC,H}$, $D_{AC,U}$, $D_{AC,L}$.\par
\emph{Generator:}
Specifically, $G_{A}$ consists of three encoders and two decoders. The first encoder learns to encode the target foreground appearance features with $X_A^1\vert^t_{t-2}=X_{TFG}\vert^t_{t-2}$ as its input. The second encoder learns to encode the source motion features with $X_A^2\vert^t_{t-2}=[X_{SP}\vert^t_{t-2},\tilde{Y}_{LO}\vert^t_{t-2}]$ as its input. The third encoder learns to encode features for previously generated foregrounds $\tilde{Y}_{FG}\vert^{t-1}_{t-2}$. Then the three kinds of features are summed and fed into the first decoder to generate $\tilde{Y}_{FG}^t$, which is the desired foreground appearance at time $t$. Meanwhile, features of the second and the third encoders are summed and fed into the second decoder to generate the shadow map $\tilde{Y}_{M}^t$, which is output by a sigmoid layer into the same size as the input background image $X_{TBG}$. Thus, the second decoder, the second and the third encoders form our \textbf{shadow extraction module}, which is light-weight because it only requires one additional decoder on the basis of foreground synthesis modules. By multiplying $X_{TBG}$ with $\tilde{Y}_{M}^t$, the background brightness is modulated pixel by pixel to achieve shadow rendering. Since the generation has no relation to background appearance, the shadow map $\tilde{Y}_{M}^t$ is identical to any $X_{TBG}$ and therefore supports shadow rendering for arbitrary images, which enables background appearance control. Then the synthetic foreground and the rendered background are added together to compose the full image $\tilde{Y}_{I}^t$, which is the desired video scene at time $t$.\par
\emph{Discriminators:}
In addition, we design multiple multi-scale discriminators ($D_{H}$, $D_{U}$, $D_{L}$, $D_{AC,H}$, $D_{AC,U}$, $D_{AC,L}$ and $D_S$) for the three part-specific ACGAN losses and one scene GAN loss.
Specifically, each part-specific \textbf{ACGAN loss} is used for the supervision of a specific body part and is made of a standard GAN loss and an appearance consistency loss, aiming at supervising visual quality and appearance consistency respectively.
\emph{As for the visual quality}, we decompose the generated and the real foreground appearances into the three body parts and feed them as the fake and the real samples into their corresponding standard body part discriminators $D_{H}$, $D_{U}$ and $D_{L}$, forcing the generator $G_A$ to synthesize more realistic body part appearances.
\emph{As for the appearance consistency}, we further apply three appearance-consistency (AC) body part discriminators $D_{AC,H}$, $D_{AC,U}$ and $D_{AC,L}$ to ensure appearances of the generated body parts are consistent with their input appearance conditions. Specifically, we obtain three kinds of body part appearance pairs as training samples for each $D_{AC}$ as shown in Figure \ref{fig4}(b): 1) consistent pair $P_1$: two body parts from the same person, labeled as "true"; 2) inconsistent pair $P_2$: two body parts from different persons, labeled as "false"; 3) fake pair $P_{fake}$: body part of the generated $\tilde{Y}_{FG}$ and the corresponding part of the input appearance condition ${X}_{TFG}$, labeled as "false" when updating discriminator and labeled as "true" when updating generator. In company with the progress of $D_{AC}$s that distinguish inconsistent body part appearances well, $G_A$ learns to generate more consistent body part appearances during adversarial training.
The \textbf{scene GAN loss} is designed to force the appearance generator $G_A$ to focus on details at part boundaries and compose the full scene harmoniously, where we feed $\tilde{Y}_{I}$ and ${Y}_{I}$ as the fake and the real samples to the scene discriminator $D_S$ for training.\par
\emph{Objective Function:}
To train the appearance GAN, we design the objective like this:
\begin{small}
\begin{equation}
\begin{aligned}
L_{A}=&L_{ACGAN}^{H}+L_{ACGAN}^{U}+L_{ACGAN}^{L}+L_{GAN}^{S}\\&+\lambda_{T}L_{T}^{A}+\lambda_{FM}L_{FM}^{A}+\lambda_{VGG}L_{VGG}^A
\label{con:E7}
\end{aligned}
\end{equation}
\end{small}$L_{ACGAN}^{H/U/L}$ are ACGAN losses of different body parts, each of which is summed by a standard GAN loss $L_{GAN}$ and an appearance-consistency loss $L_{AC}$. Since all of them have the same design, we only give the derivation of $L_{ACGAN}^{H}$ as an example:
\begin{small}
\begin{equation}
\begin{aligned}
L_{ACGAN}^{H}=&L_{GAN}^H+\lambda_{AC}L_{AC}^H
\label{con:E8}
\end{aligned}
\end{equation}
\end{small}
\begin{small}
\begin{equation}
\begin{aligned}
L_{GAN}^H=&E[log D_{H}(Y_{FG,H},X_{A,H})\\&+log[1-D_{H}(\tilde{Y}_{FG,H},X_{A,H})]]
\label{con:E9}
\end{aligned}
\end{equation}
\end{small}
\begin{small}
\begin{equation}
\begin{aligned}
L_{AC}^H=&E[log D_{AC,H}(P_{1,H})\\&+log[1- D_{AC,H}(P_{2,H})]\\&+log[1-D_{AC,H}(P_{fake,H})]]
\label{con:E10}
\end{aligned}
\end{equation}
\end{small}where $\lambda_{AC}$ is the weight of $L_{AC}$, $Y_{FG,H}$ represents head region of the real foreground, $X_{A,H}$ represents the conditioning input obtained by concatenating head regions of $X_A^1$ and $X_A^2$, $P_{1,H}$, $P_{2,H}$ and $P_{fake,H}$ are consistent, inconsistent and fake head appearance pairs respectively.\\
$L_{GAN}^{S}$ is the scene GAN loss, derived as follows:
\begin{small}
\begin{equation}
\begin{aligned}
L_{GAN}^{S}=&E[log D_{S}(Y_{I},X_{I})+log[1-D_{S}(\tilde{Y}_{I},X_{I})]]
\label{con:E11}
\end{aligned}
\end{equation}
\end{small}where $Y_{I}$ is the real scene image with respect to $\tilde{Y}_{I}$, $X_I=[X_A^1,X_A^2,X_{TBG}]$.\\
$L_{T}^{A}$ is the temporal loss weighted by $\lambda_{T}$ to improve temporal consistency, which can be derived as follows:
\begin{small}
	\begin{equation}
	\begin{aligned}
	L_{T}^{A}=E[log D_{A}^{T}({S_{I}})+log[1-D_{A}^{T}(\tilde{S}_{I})]]
	\label{con:E12}
	\end{aligned}
	\end{equation}
\end{small}where $D_{A}^{T}$ is the temporal discriminator of the appearance GAN, trained to determine whether an image sequence is real or fake. $S_{I}$ and $\tilde{S}_{I}$ are the real and the generated image sequences, which are obtained similarly to the layout sequences by concatenating three consecutive ${Y}_{I}$s and $\tilde{Y}_{I}$s.\\
$L_{FM}^{A}$ is the discriminator feature matching loss weighted by $\lambda_{FM}$, $L_{VGG}^A$ is the VGG loss \cite{johnson2016perceptual,dosovitskiy2016generating,wang2018high} weighted by $\lambda_{VGG}$.

%% file: experiments.tex
\section{Experiments}
\label{experiments}
\subsection{Solo Dance Dataset}
\label{dataset}
We construct a large solo dance dataset with 124 dance videos collected from 58 males and 66 females, including a variety of human identities and clothing styles that allow for appearance generalization. Our dataset covers four main dance types (modern, jazz, rumba, tap) with the dancer in each video performing a dance different from others. Each video is an individual dance clip captured at a 30fps frame rate, containing 1200 continuous frames with the corresponding appearances and motions. To satisfy the setting that single persons perform difficult movements in stationary backgrounds, only solo dance videos with fixed viewpoints are included in the dataset.\par
After the videos are collected, we automatically extract backgrounds for each video by stitching detached background regions of different frames. Then we detect poses, layouts and foregrounds for each frame, where we crop and resize all the frames to central 192x256 regions and manually rectify ones with bad detection results for better data quality. Next, we divide each processed video sequence into two halves, where the first half is used to extract $X_{SP}$s and the second is used to obtain the paired $X_T$s. Therefore, we obtain 600 available conditioning inputs and the corresponding ground-truth frames for each of the 124 videos. In our experiments, we use 100 videos for training and the remaining 24 videos for testing.
\begin{figure*}
\centerline{\includegraphics[width=0.98\textwidth]{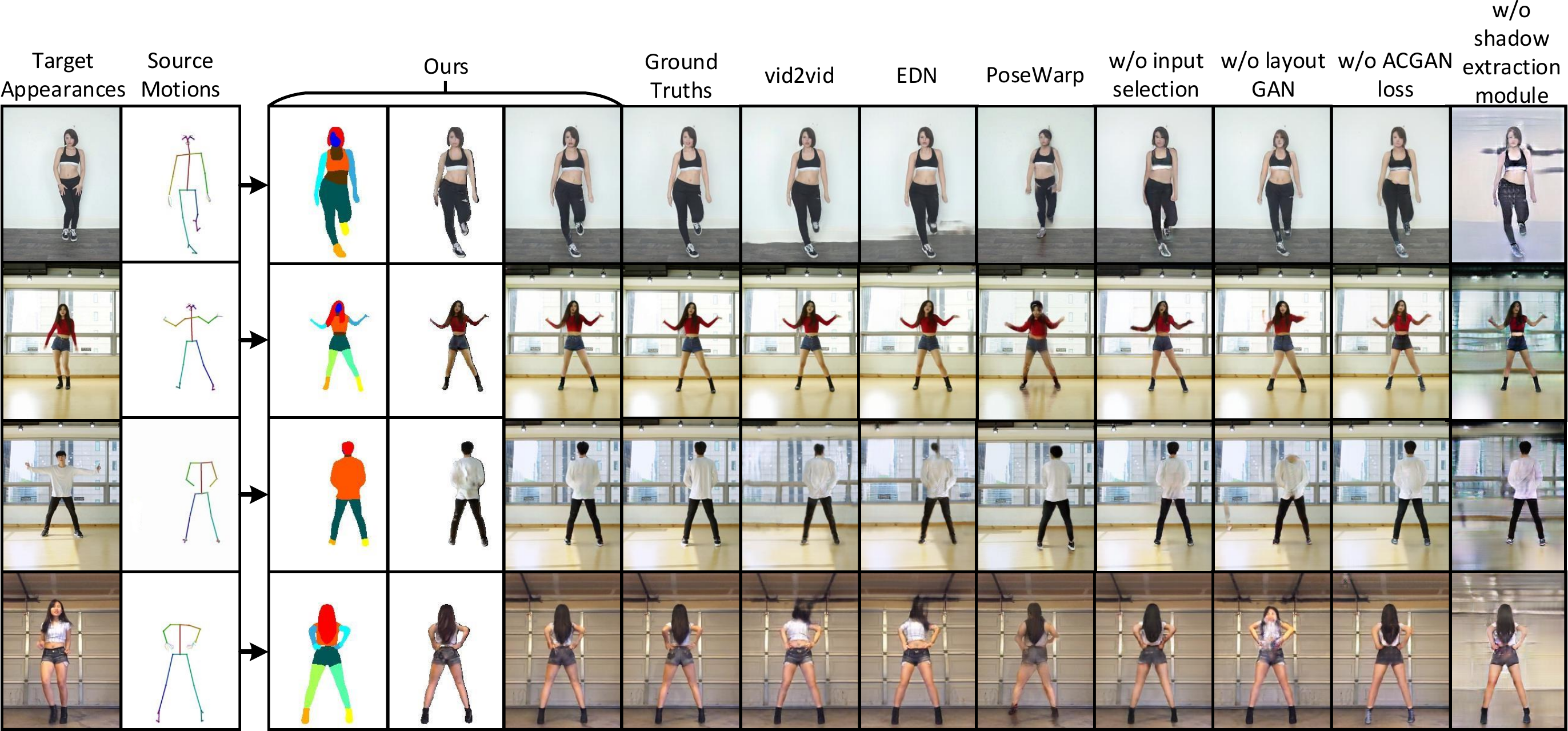}}
\caption{Qualitative comparison results on HVMT tasks (please zoom in for a better view). From left to right: input target appearances, input source motions, our generated results (layouts, foregrounds, full scenes), ground-truth frames, results of vid2vid \cite{wang2018video}, results of EDN \cite{chan2019everybody}, results of PoseWarp \cite{balakrishnan2018synthesizing}, results of the four ablated variants with respect to input selection strategy, layout GAN, ACGAN loss and shadow extraction module.} \label{fig5}
\end{figure*}
\begin{figure*}
\centerline{\includegraphics[width=0.98\textwidth]{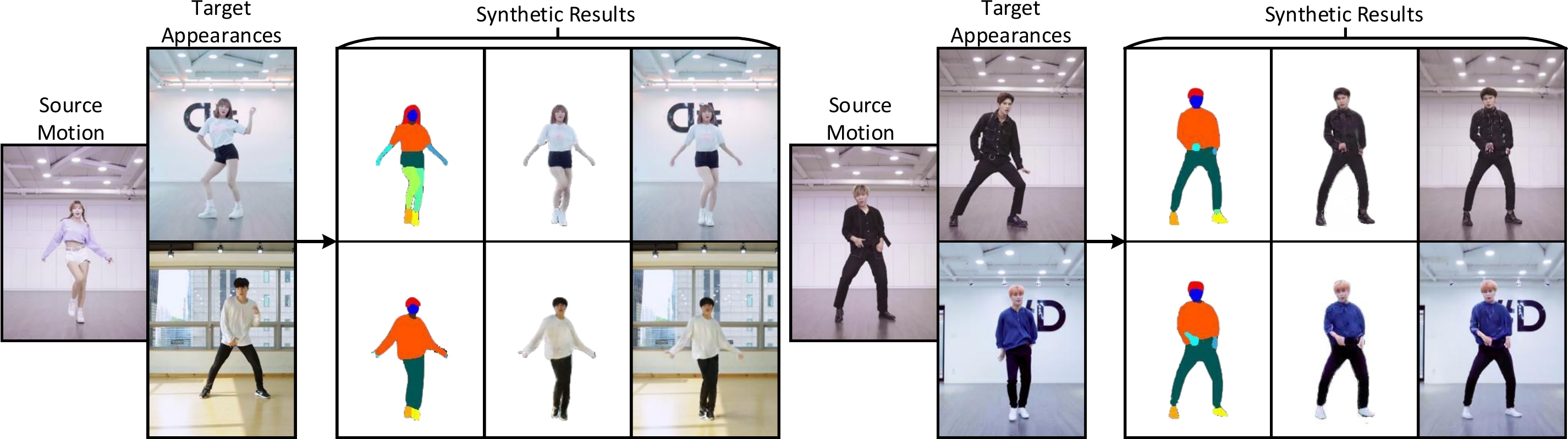}}
\caption{Examples of ordinary HVMT (please zoom in for a better view). Each synthetic result (layout, foreground, full scene) is generated to have the same motion as its input source motion image and the same appearance as its input target appearance image.} \label{fig6}
\end{figure*}
\subsection{Experimental Setup}
\subsubsection{Our Method}
The design of encoders and decoders follows pix2pixHD \cite{wang2018high}, where the numbers of convolutional filters are decreased to half of the original pix2pixHD to reduce the model size. All the discriminators that distinguish single frames (standard and AC discriminators) follow the multi-scale PatchGAN architecture \cite{isola2017image}, and each of them has three spatial scales to model different image resolutions. All the temporal discriminators that distinguish sequences rather than single frames follow the design of \cite{wang2018video}, and each of them has three time scales to ensure both short-term and long-term temporal consistency.\par
During the training stage, the layout GAN and the appearance GAN are trained separately with Adam optimizers (learning rate: 0.0002, batch size: 4) on 4 Nvidia RTX 2080 Ti GPUs for 10 epochs, where we set $\lambda_{AC}=5$ and $\lambda_{SS}=\lambda_T=\lambda_{FM}=\lambda_{VGG}=10$ in the objective functions. Since frames (layouts and foregrounds) at time -1 and -2 don't exist, we directly replace them with two same-size zero tensors to first generate the frame at time 0, which is then taken as input together with the zero tensor at time -1 to generate the frame at time 1. By doing this during training, the layout GAN and the appearance GAN learn to handle the generation of the first two frames.
%可以加上运行时间？
\begin{figure*}
\centerline{\includegraphics[width=0.78\textwidth]{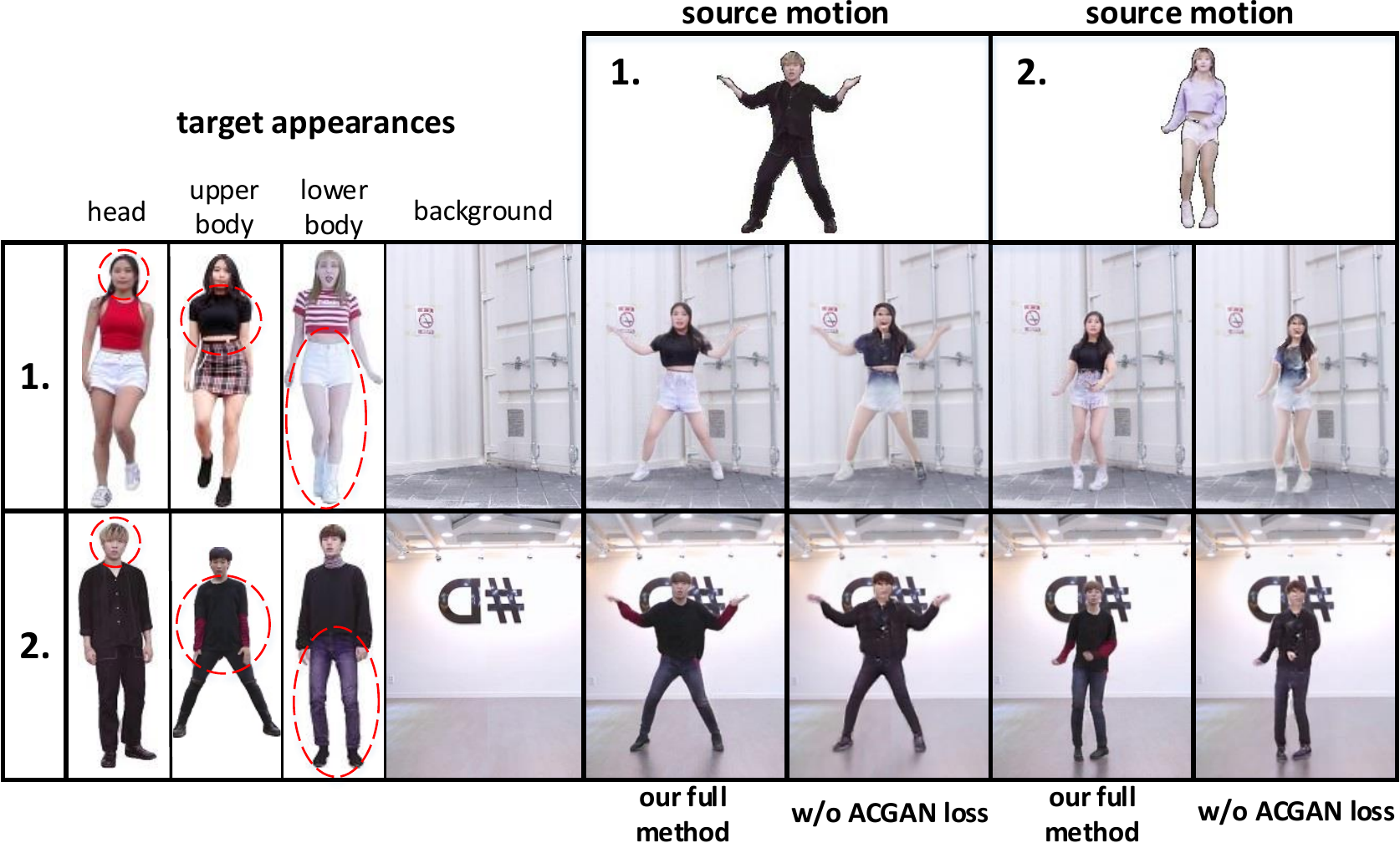}}
\caption{Examples of multi-source appearance control (please zoom in for a better view). Each synthetic image is generated based on five inputs in terms of body motion, appearances of head, upper body, lower body and background. We also show the results generated by the variant model without ACGAN loss, allowing for comparisons with our full method.} \label{fig7}
\end{figure*}
\subsubsection{Other Methods}
We also implement the following methods for comparisons:
\begin{itemize}
\item{\textbf{Video-based methods}:\\We compare our method with two state-of-art video-based methods vid2vid \cite{wang2018video} and EDN \cite{chan2019everybody}, both are video-specific with each model can only generate videos with the same scene appearance. In our implementation, each of their models is trained with 3000 frames of one specific video.}
\item{\textbf{Image-based methods}:\\Since video-based methods are video-specific, we implement a state-of-art image-based method PoseWarp \cite{balakrishnan2018synthesizing} as a general-purpose baseline, which is trained on the same data as ours in a general way.}
\item{\textbf{w/o input selection}:\\To evaluate the effectiveness of our input selection strategy which enables input appearance control, we implement a model trained with body part appearance conditions selected randomly with no extra computation.}
\item{\textbf{w/o layout GAN}:\\To evaluate the effectiveness of our layout GAN that provides more accurate motion conditions, we implement a model with only the appearance GAN, which is fed with only 2D poses as motion conditions.}
\item{\textbf{w/o ACGAN loss}:\\To evaluate the effectiveness of our ACGAN loss which enables foreground appearance control, we implement a model whose appearance GAN is trained without ACGAN loss.}
\item{\textbf{w/o shadow extraction module}:\\To evaluate the effectiveness of our shadow extraction module which enables background appearance control, we implement a model that generates backgrounds from scratch with fixed background images included in its input appearance conditions.}
\end{itemize}
\subsection{Qualitative Results}
To assess the quality of our synthetic results, we test different methods on our test set and compare their synthetic frames with ground-truth video frames. It's noted that ground truths are available here because the motion and the appearance conditions of each synthetic frame are obtained from the same person as has been stated in the description of our dataset (Sec.\ref{dataset}). As shown in Figure \ref{fig5}, we randomly visualize some synthetic frames generated by our method and other methods to make qualitative comparisons. Based on the proposed GAC-GAN, we can synthesize motion transfer videos with realistic appearance and body pose details, which are consistent with the input target appearances and source motions. In contrast, the image-based method PoseWarp \cite{balakrishnan2018synthesizing} can't preserve the target appearances well with body poses and locations are not consistent with the desired source motions. Although the two video-based methods vid2vid \cite{wang2018video} and EDN \cite{chan2019everybody} perform well when synthesizing appearances of frequent poses (e.g., front bodies in the first two rows of Figure \ref{fig5}), they render bad visual results when synthesizing appearances of infrequent poses (e.g., backside bodies in the last two rows of Figure \ref{fig5}). We think the main reason is that infrequent poses are less explored during training due to the imbalance between numbers of frequent and infrequent poses in their training data, which contains only one video sequence for each video-specific model.
However, the quality of our results is not influenced by such imbalance because we provide our model with optimal appearance inputs that contain the maximum texture information needed for appearance synthesis. Besides, our model is trained with access to more infrequent poses contained in the whole dataset, leading to better results than EDN and vid2vid when synthesizing unseen infrequent pose appearances.
Please refer to our supplementary material for the video version of the qualitative comparison results.\par
%The video version of qualitative comparison results are available \href{https://youtu.be/ijUapNfipbc}{online}.\par
Then we test our method on tasks which have no ground-truth frame for a better understanding of the appearance-controllable human video motion transfer that we realize:
\begin{figure*}
\centerline{\includegraphics[width=0.75\textwidth]{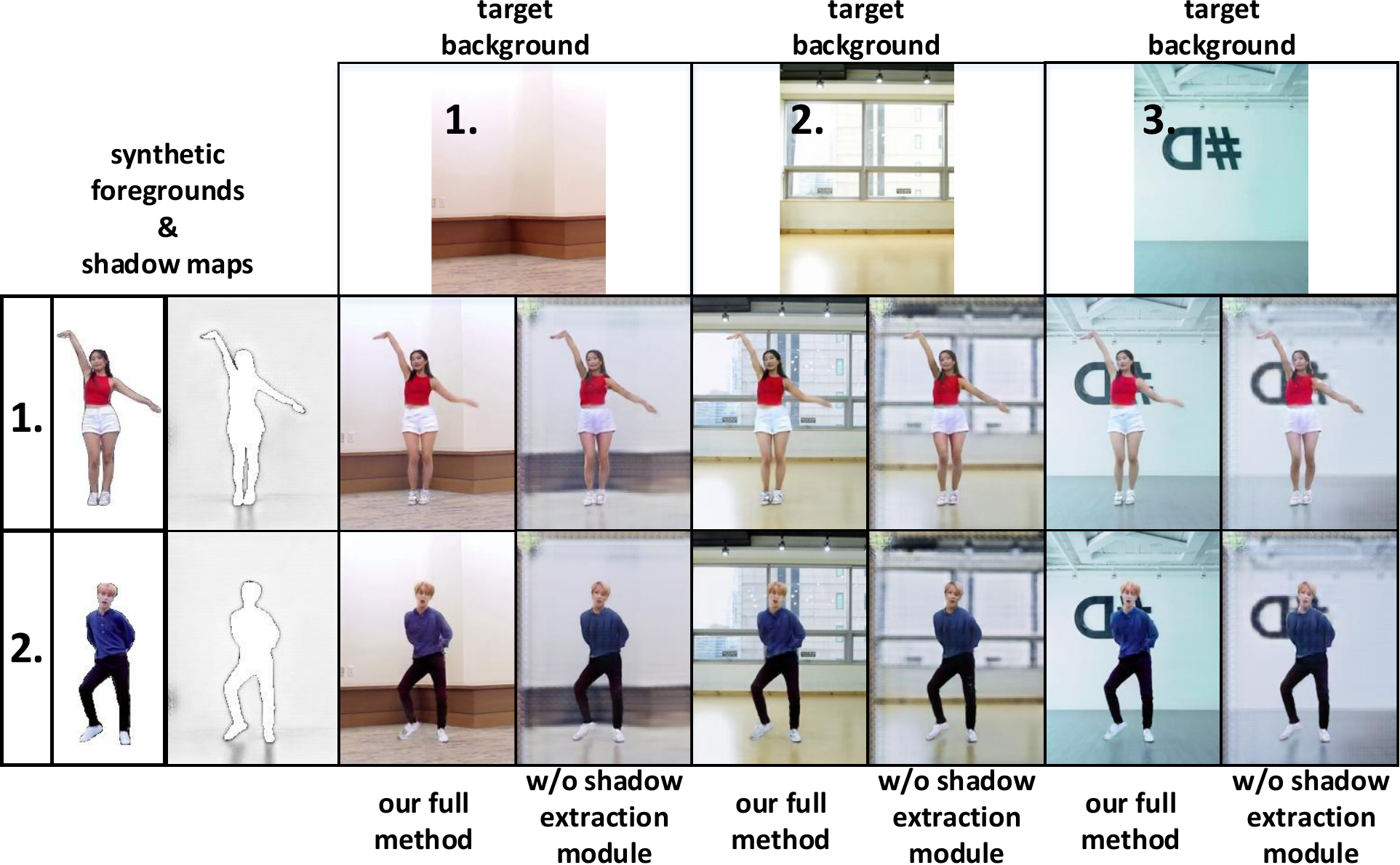}}
\caption{Examples of background appearance control (please zoom in for a better view). The input backgrounds are modulated by the synthetic shadow maps to fuse with the synthetic foregrounds. We also show the results generated by the variant model without shadow extraction module, allowing for comparisons with our full method.} \label{fig8}
\end{figure*}
\begin{itemize}
\item{\textbf{Ordinary HVMT}:\\Transfer one person's motion to another person without further appearance control on body part foregrounds and surrounding backgrounds, which has the same test setting as the qualitative comparisons shown in Figure \ref{fig5} except for the absence of ground-truth frames. As shown in Figure \ref{fig6}, our one-time trained model can generate high-quality motion transfer video frames with details of the source motions and the target appearances are well preserved.}
\item{\textbf{Appearance-Controllable HVMT}:\\For HVMT with multi-source appearance control, we let our model synthesize videos with appearances of body parts and backgrounds coming from different appearance sources, where the synthetic appearances are naturally composed and the synthetic motions are consistent with the source motions as shown in Figure \ref{fig7}. For further evaluation on background appearance control based on our synthetic shadow maps, we add shadows to different backgrounds and fuse them with the synthetic foregrounds to achieve controllable background replacement, where detailed shadows are rendered in harmony with the human motions as shown in Figure \ref{fig8}.}
\end{itemize}
\par
%For a full and animated version of our synthetic visual results, please refer to our \href{https://github.com/wswdx/Appearance-Composing-GAN/}{online video examples}.
For a full and animated version of the synthetic visual results, please refer to our supplementary materials.
%For a full and animated version of our synthetic visual results, please refer to our \href{https://youtu.be/8fhr5bcFM6Y}{online video examples}.
\begin{table*}
\begin{center}
{\caption{Quantitative and perceptual comparison results. SSIM and PSNR are similarity metrics, the higher the better. LPIPS and VFID are distance metrics, the lower the better. Preference score is denoted as the proportion of perceptually preferred videos generated by our method.}\label{table1}}
\resizebox{\textwidth}{!}{
\begin{tabular}{lccc"cccc"c}
  \thickhline
  & vid2vid \cite{wang2018video} & EDN \cite{chan2019everybody} & PoseWarp \cite{balakrishnan2018synthesizing} & \tabincell{c}{w/o\\input selection} & \tabincell{c}{w/o\\layout GAN} & \tabincell{c}{w/o\\ACGAN loss} & \tabincell{c}{w/o\\shadow extraction module} & Ours\\
  \hline 
  SSIM & 0.8834 & 0.8711 & 0.8380 & 0.8652 & 0.8545 & 0.8613 & 0.8580 & \textbf{0.8947}\\
  PSNR & 26.9316 & 26.5653 & 23.5423 & 24.7923 & 22.5398 & 24.1372 & 24.0782 & \textbf{27.9458}\\
  LPIPS & 0.0352 & 0.0363 & 0.0537 & 0.0413 & 0.0436 & 0.0394 & 0.0419 & \textbf{0.0341}\\
  VFID & 3.9752 & 4.3410 & 7.0721 & 5.1426 & 5.3624 & 4.9845 & 5.2187 & \textbf{3.9689}\\
  Preference Score & 69.2\% & 73.1\% & 93.8\% & 76.2\% & 81.5\% & 77.7\% & 80.8\% & ---\\
  \thickhline
\end{tabular}}
\end{center}
\end{table*}
\subsection{Quantitative Results}
We also make a quantitative assessment to analyze differences between synthetic and ground-truth video frames by four metrics: Structural Similarity (SSIM), Peak Signal to Noise Ratio (PNSR), Learned Perceptual Image Patch Similarity (LPIPS) \cite{zhang2018unreasonable} and Video Fr{\'e}chet Inception Distance (VFID) \cite{wang2018video}. In particular, SSIM and PSNR are classic metrics that measure the pixel-level image similarity between synthetic results and ground truths, which are simple and based on shallow functions. LPIPS is a newly invented metric that accounts for similarity measurement between two images, which is computed based on features extracted by deep models. VFID is also a deep metric with a video recognition CNN model performing as its feature extractor, which measures temporal consistency in addition to visual quality. It's noted that SSIM and PSNR are similarity metrics while LPIPS and VFID are distance metrics, which means higher values are better for the former while the opposite for the latter. All the comparison results are summarized in the first five rows of Table \ref{table1}. We can see that our method outperforms other methods for all the metrics, which indicates that our synthetic results have not only higher visual quality but also better temporal consistency.
\subsection{Human Perceptual Results}
For human perceptual assessment, we conduct a human subjective study by performing preference tests on the Amazon Mechanical Turk (AMT). Particularly, each question is an A/B test where we show turkers two videos generated by our method and a compared method and let them choose which video looks more realistic in consideration of visual quality and temporal consistency. After gathering 10 answers for 13 videos generated by different methods, we summarize the average human preference scores in the last row of Table \ref{table1}. The results indicate that videos generated by our method are also perceptually preferred to those generated by others, which is consistent with our qualitative and quantitative results. It's noted that each preference score in the table represents the proportion of perceptually preferred videos generated by our full method when compared to one of the compared methods. Since the comparison between our full method and itself is missing, the preference score of our full method is left blank.
\subsection{Ablation Studies}
We also compare our full method with the above mentioned four variants with respect to ablations of our input selection strategy, layout GAN, ACGAN loss and shadow extraction module. As can be seen from the quantitative and the perceptual results shown in the 5-8th columns of Table \ref{table1}, our full method outperforms all the variants significantly, which indicates that videos generated by our full method have higher visual quality and better temporal consistency than those generated by the variants without our important components.
We also make comparisons on qualitative results as shown in the last four columns of Figure \ref{fig5}. The 10th and the 12th columns indicate that, without the selected optimal appearance inputs and the elaborate ACGAN loss, the model can't preserve human appearances well, which results in blurry faces and bodies, proving that both the two components can improve human appearance details. The 11th column indicates that, without the layout GAN to provide additional motion conditioning inputs that describe human motions more accurately, the model even can't generate the desired body motions, let alone satisfactory appearances, which proves the effectiveness of our two-stage framework design. The last column shows that, without shadow rendering, the model fails in both background and foreground synthesis, which proves that our shadow extraction module can also improve foreground synthesis quality.
Moreover, we make further qualitative comparisons to demonstrate the importance of our ACGAN loss and shadow extraction module in the multi-source foreground and background appearance control respectively. As shown in Figure \ref{fig7}, without training with the proposed ACGAN loss, the model renders bad body part appearances which are mixed up together and inconsistent with the input appearance conditions. As shown in Figure \ref{fig8}, without the synthetic shadow maps to achieve shadow rendering, the model can only generate background appearances from scratch, which results in blurry backgrounds as well as unrealistic foregrounds.

%% file: conclusion.tex
\section{Conclusion}
\label{conclusion}
In this paper, we present GAC-GAN for general-purpose and appearance-controllable human video motion transfer. To synthesize videos with controllable appearances, we propose a multi-source input selection strategy to first obtain controllable input appearance conditions. Moreover, given such appearance-controllable inputs, we propose a two-stage GAN framework trained with the ACGAN loss and implanted with the shadow extraction module to enable the compatible synthesis of the appearance-controllable outputs. Extensive experiments on our large-scale solo dance dataset show that our proposed method can not only enable appearance control in a general way but also achieve higher video quality than state-of-art methods. We also conduct comprehensive ablation studies with respect to our input selection strategy, layout GAN, ACGAN loss and shadow extraction module. The results show that our full method achieves higher performance than all the ablated variants, which proves the effectiveness of our important components. Although our method performs well in most cases, challenges and open problems remain: 1) Since the GAC-GAN is a deep learning model, our method may fail when the GAN model is tested on unseen domains that are too different from the training domains (e.g., generate videos for CG characters rather than real humans). 2) Because the quality of the synthetic outputs highly depends on the conditioning inputs which are obtained based on pose and layout estimation techniques, texture artifacts may occur when these estimations fail. In the future, we may also explore the potential of synthesizing more complex videos where multiple people dance together rather than solo dance videos. Besides, video synthesis with movable camera views is also worth studying, requiring further consideration of background motions. Both of them are promising extensions to our accomplished work.